\documentclass[preprint,12pt]{elsarticle}

\usepackage{amsthm, amsmath, amssymb, bm}
\usepackage{graphicx}
\usepackage{booktabs}
\usepackage{multirow}
\usepackage{subcaption}
\usepackage{algorithm2e}
\usepackage{natbib}
\usepackage[colorlinks=true,
            linkcolor=blue,
            urlcolor=blue]{hyperref}
\journal{Structural Safety}

\theoremstyle{plain}

\DeclareMathOperator*{\argmax}{arg\,max}
\DeclareMathOperator*{\argmin}{arg\,min}

\begin{document}

\begin{frontmatter}

\title{AK-MCS-C$^2$: Active Kriging Monte Carlo Simulation method with conformal certification for failure probability estimation}

\author[ENS]{Edgar Jaber}
\author[EDF]{Vincent Chabridon}
\author[ENS]{Mathilde Mougeot}

\affiliation[ENS]{organization={Université Paris-Saclay, CNRS, ENS Paris-Saclay, Centre Borelli},
            city={Gif-sur-Yvette},
            postcode={91190},
            country={France}}
\affiliation[EDF]{organization={EDF R\&D},
            addressline={6 Quai Watier},
            city={Chatou},
            postcode={78401},
            country={France}}
\begin{abstract}
We introduce a novel active-learning framework for failure probability estimation in structural reliability analysis that integrates active Kriging Monte Carlo simulation with conformal prediction. The proposed approach employs an adaptive cross-conformal strategy specifically designed for small-sample settings and kriging surrogate models using the J+GP conformal estimator. Unlike standard AK-MCS methods, the proposed framework provides distribution-free guarantees on prediction errors, leading to more reliable classification of samples near the limit-state surface. This improved uncertainty quantification enhances both the accuracy and robustness of failure probability estimates, especially for rare-event regimes where such efficiency is crucial. Reproducible numerical results illustrate the effectiveness of the method and also compare it to classical approaches on well-established benchmarks.
\end{abstract}

\begin{highlights}
\item We propose a conformalized Active Kriging Monte Carlo Simulation (AK-MCS-C$^2$) framework for structural reliability analysis.
\item The method integrates Gaussian process (GP) surrogates with cross-conformal prediction to obtain distribution-free uncertainty quantification of the surrogate at every Monte Carlo point.
\item The proposed J+GP conformal estimator yields adaptive prediction intervals that define a principled learning function for active enrichment.
\item The procedure provides certified two-sided bounds on the failure probability estimate.
\item Numerical experiments demonstrate improved robustness and accuracy in rare-event probability estimation, and the paper is accompanied by reproducible source code.
\end{highlights}


\end{frontmatter}

\section{Introduction}
\label{sec1}
\noindent Structural reliability analysis aims at quantifying the probability of failure of engineering systems under uncertainty. In a probabilistic setting, the system response is described by a limit-state function $g(\bm{X})$, where $\bm{X}\in\mathcal{X}\subseteq\mathbb{R}^d$ denotes a vector of uncertain input parameters following a distribution $\mathbb{P}_{\bm{X}}$ for a simulation code $g$. Failure is defined by the event $\mathcal{F} = \{\omega\in\Omega, \; g(\bm{X}(\omega)) \leq 0\}$ \citep{Ditlevsen1996,Lemaire2009,Melchers2018}. The associated failure probability is therefore expressed as a high-dimensional integral:
\begin{equation}
    P_{f} = \int_{\mathcal{X}}\bm{1}\{g(\bm{x}) \leq 0\}d\mathbb{P}_{\bm{X}}(\bm{x})
    \label{eq:p_f}
\end{equation}
This integral is rarely tractable analytically and must therefore
be approximated numerically. Classical approaches include approximation methods such as the First- and Second-Order Reliability Methods (FORM/SORM) \citep{Hasofer1974,Rackwitz1978,Breitung1984}, as well as simulation-based techniques such as Monte Carlo simulation, importance sampling \citep{Melchers1989, Chabridon2018, Fekhari2024}, and subset simulation \citep{Au2001}. While simulation methods are robust and general, their computational cost becomes prohibitive in the presence of expensive models and rare failure events. To alleviate this issue, surrogate-based methods have emerged as a powerful alternative. In particular, active-learning strategies iteratively construct surrogate models that focus computational effort near the limit-state surface. Among these approaches, Gaussian process (GP) models have played a central role due to their ability to provide both predictions and uncertainty estimates. Landmark contributions include the Efficient Global Reliability Analysis method \citep{Bichon2008} and, more prominently, the Active Kriging Monte Carlo Simulation (AK-MCS) method introduced by \citep{Echard2011}. The latter combines a Kriging surrogate with Monte Carlo sampling and an adaptive enrichment strategy, and has become a cornerstone of modern reliability analysis \citep{Moustapha2022}. Numerous extensions have been proposed, including variants based on importance sampling \citep{Echard2013,Dubourg2013}, subset simulation \citep{Huang2016}, and alternative surrogate models \citep{Marelli2018,Schobi2017}. Recent surveys highlight that many of these developments can be interpreted within a unified framework involving a surrogate model, a reliability estimator, a learning function, and a stopping criterion \citep{Teixeira2021,Moustapha2022}. Despite their success, AK-MCS-type methods rely on uncertainty measures that are inherently model-dependent. In particular, the Kriging predictive variance reflects uncertainty under the Gaussian-process assumptions and estimated hyperparameters, but does not provide distribution-free guarantees on the true prediction error \citep{Jaber2025}. Since the failure probability depends critically on the correct classification of samples near the limit-state surface, misclassification errors may significantly impact the accuracy of the estimate, especially in rare-event regimes. Furthermore, stopping criteria in active-learning reliability are often heuristic or conservative, and their statistical interpretation remains limited \citep{Moustapha2022}.\\

\noindent In parallel, conformal prediction (CP) has emerged as a powerful framework for distribution-free uncertainty quantification in machine learning \citep{Vovk2005, Shafer2008, Romano2019, Angelopoulos2021, Barber2021}. Under the assumption of exchangeability (of which independently and identically distributed is a particular case), conformal methods provide finite-sample coverage guarantees on predictive uncertainty, independently of the underlying model. For a certain machine learning model $\widehat{g}$ with inputs $\bm{X}$ learned on a dataset of size $n$, this translates to building prediction intervals $\widehat{C}_{n,\alpha}(\bm{X})$ such that the following marginal coverage property holds:
\begin{equation}
    \mathbb{P}_{\underline{\bm{Z}}_{n+1}}(g(\bm{X}^{(n+1)})\in\widehat{C}_{n,\alpha}(\bm{X}^{(n+1)}))\geq 1-\alpha,
\end{equation}
where $\bm{X}^{(n+1)}$ is a new point not seen by the training sample and the probability is taken with respect to the training design of experiments and \emph{added test point} $\underline{\bm{Z}}_{n+1} = (\bm{X}^{(i)},g(\bm{X}^{(i)}))_{i=1}^{n+1}$. In the field of computer experiments, such approaches have recently been explored in \citep{Jaber2025}. A central challenge in CP is the construction of adaptive prediction intervals whose width varies according to the local prediction error of the metamodel. Adaptivity is closely linked to the expressivity of the surrogate model meaning that intervals should be narrow where the model is accurate and wider where the error is higher. This feature is still challenging to obtain for many surrogates \citep{Hatstatt2026}. However for GPs, this has recently been obtained for cross-conformal estimators by \citep{Jaber2025} by extending classical Jackknife+ estimators \citep{Barber2021,Vovk2015}.\\

\noindent This work aims at bridging these two lines of research by introducing a conformal certification layer within the AK-MCS framework. The proposed method augments the surrogate-based active-learning procedure with conformal calibration, yielding distribution-free uncertainty quantification for the GP surrogate predictions. This calibrated uncertainty is then propagated to the classification of failure events and to the estimation of the failure probability, enabling the construction of certified bounds for $P_f$ as well as robust stopping criteria and empirical coverage diagnostic. The main contributions of this paper are threefold. First, we develop a conformalized active-learning strategy tailored to reliability analysis with Kriging surrogates. Second, we derive certified failure-probability estimators based on conformal prediction sets. Third, we assess the performance of the proposed method on benchmark problems, with a focus on accuracy, efficiency, and calibration properties in rare-event regimes and we provide a reproducible code in the following GitHub repository : \href{https://github.com/EdgarJaber/AK-MCS-C2}{\texttt{EdgarJaber/AK-MCS-C2}}.\\

\noindent The rest of this paper is structured as follows. Section~\ref{sec2} introduces the notations used throughout the manuscript, section~\ref{sec3} presents the mathematical theory underlying the proposed methodology: section~\ref{sec31} recalls the Kriging surrogate modeling approaches, section~\ref{sec32} introduces the J+GP cross-conformal estimator, and section~\ref{sec33} describes the AK-MCS framework together with the different learning functions, including the proposed conformal C$^2$ strategy in section~\ref{sec333}. The numerical results are reported in section~\ref{sec4}, where the performance of the method is assessed on benchmark reliability problems. Finally, section~\ref{sec5} concludes the paper and discusses perspectives for future work.

\section{Notations}
\label{sec2}
\noindent Let $(\Omega,\mathcal{F},\mathbb{P})$ a probability space, $\mathcal{X} \subset \mathbb{R}^d$ denote the input space and let $\bm{X} \sim \mathbb{P}_{\bm{X}}$ be a random vector with probability measure $\mathbb{P}_{\bm{X}} = \mathbb{P}(\bm{X}^{-1}(\{.\}))$. The performance (or limit-state) function is denoted by $g : \mathcal{X} \rightarrow \mathbb{R}$, and the failure domain is defined as $\mathcal{F} = \{\bm{x} \in \mathcal{X} : g(\bm{x}) \leq 0\}$. The associated probability of failure is given by:
\begin{equation}
P_f = \mathbb{P}(g(\bm{X}) \leq 0) = \int_{\mathcal{X}} \mathbf{1}\{g(\bm{x}) \leq 0\} \, \mathrm{d}\mathbb{P}_{\bm{X}}(\bm{x}).
\end{equation}
We denote by $\mathrm{DoE}_{n} = \{(\bm{x}^{(i)}, g(\bm{x}^{(i)}))\}_{i=1}^n$ the design of experiments of size $n$. A Gaussian process surrogate model trained on $\mathrm{DoE}_{n}$ provides a posterior mean $\widetilde{g}(\bm{x})$ and a posterior standard deviation $\widetilde{\sigma}(\bm{x})$ at any point $\bm{x} \in \mathcal{X}$. The leave-one-out (LOO) predictors are denoted by $\widetilde{g}_{-i}(\bm{x})$ and $\widetilde{\sigma}_{-i}(\bm{x})$. A Monte Carlo sample is denoted by $\mathcal{D} = \{\bm{x}^{(i)}\}_{i=1}^N$, where $N$ is the sample size. The indicator function is written as $\mathbf{1}\{\cdot\}$. Throughout the paper, $\varepsilon > 0$ denotes a small regularization constant used to avoid numerical instabilities. For conformal prediction, we consider a miscoverage level $\alpha \in (0,1)$ and denote by $\widehat{q}_{n,\alpha}^{\pm}\{\cdot\}$ the empirical quantiles used to construct prediction intervals. For any finite subset $\{v_{i}\}_{i=1,\ldots,n}$ of an ordered set, the $(1-\alpha)$-empirical quantile, with $\alpha \in (0, 1)$, is given by:
\begin{equation}
    \widehat{q}^{\;+}_{n,\alpha}\left\{ v_{i} \right\}:= \mathrm{the}\;\lceil (1-\alpha)(n+1) \rceil \text{-th smallest value of} \; v_{1},\ldots,v_{n}\;,
\end{equation}
with $\lceil \cdot \rceil$ denotes the ceil function. Similarly, the $\alpha$-empirical-quantile is given by:
\begin{equation}
    \widehat{q}^{\;-}_{n,\alpha}\left\{ v_{i} \right\}:= \mathrm{the}\;\lfloor \alpha(n+1) \rfloor \text{-th smallest value of} \; v_{1},\ldots,v_{n} = -\widehat{q}^{\;+}_{n,\alpha}\left\{ - v_{i} \right\}\;,
\end{equation}
where $\lfloor \cdot \rfloor$ denotes the floor function. The resulting conformal prediction sets at level $1-\alpha$ and prediction point $\bm{x}$ are denoted by $\widehat{C}_{n,\alpha}(\bm{x})$. Finally, at iteration $t$ of the active-learning procedure, the current design is denoted by $\mathcal{D}_t$ of size $n_t$, and the corresponding GP predictor by $(\widetilde{g}_t, \widetilde{\sigma}_t)$. The learning function is denoted by $\mathcal{L}_t(\bm{x})$, and the next enrichment point by $\bm{x}_{t+1}$.

\section{Methodology}
\label{sec3}
\subsection{Kriging methods}
\label{sec31}
\noindent We model the limit-state function $g$ as a realization of a Gaussian process (GP) defined on $\mathcal{X} \subset \mathbb{R}^d$. More precisely, we assume:
\begin{equation}
g(\bm{x}) = m(\bm{x}) + W(\bm{x}),
\end{equation}
where $m(\bm{x})$ is a deterministic mean function and $W(\bm{x})$ is a centered Gaussian process with covariance function:
\begin{equation}
\mathrm{Cov}(W(\bm{x}), W(\bm{x}')) = k_{\theta}(\bm{x}, \bm{x}'),
\end{equation}
parameterized by hyperparameters $\theta$. In this work, we consider a constant mean $m(\bm{x}) = \beta$ and a covariance kernel $k_{\theta}:\mathcal{X}\times\mathcal{X}\to \mathbb{R}$. Given a design of experiments $\mathrm{DoE}_{n} = \{(\bm{x}^{(i)}, g(\bm{x}^{(i)}))\}_{i=1}^n$, let $\bm{g}_n = (g(\bm{x}^{(1)}), \ldots, g(\bm{x}^{(n)}))^\top$ denote the vector of observations, and define the covariance matrix $\bm{K}_n \in \mathbb{R}^{n \times n}$ with entries:
\begin{equation}
(\bm{K}_n)_{ij} = k_{\theta}(\bm{x}^{(i)}, \bm{x}^{(j)}).
\end{equation}
For a new input $\bm{x} \in \mathcal{X}$, we define the covariance vector:
\begin{equation}
\bm{k}_n(\bm{x}) = \big(k_{\theta}(\bm{x}, \bm{x}^{(1)}), \ldots, k_{\theta}(\bm{x}, \bm{x}^{(n)})\big)^\top.
\end{equation}
Under the GP prior, the joint distribution of $(\bm{g}_n, g(\bm{x}))$ is Gaussian, and conditioning yields the posterior (Kriging) predictor. The posterior mean is given by:
\begin{equation}
\widetilde{g}(\bm{x}) = \beta + \bm{k}_n(\bm{x})^\top \bm{K}_n^{-1} (\bm{g}_n - \beta \bm{1}_{n}),
\end{equation}
and the posterior variance reads:
\begin{equation}
\widetilde{\sigma}^2(\bm{x}) := k_{\theta}(\bm{x}, \bm{x}) - \bm{k}_n(\bm{x})^\top \bm{K}_n^{-1} \bm{k}_n(\bm{x}),
\end{equation}
where $\bm{1}_{n}$ is the vector of ones in $\mathbb{R}^n$. The posterior standard deviation is then $\widetilde{\sigma}(\bm{x}) = \sqrt{\widetilde{\sigma}^2(\bm{x})}$. The hyperparameters $(\beta, \theta)$ are estimated by maximum likelihood, leading to the classical Kriging predictor \citep{Rasmussen2006}.  \\

\noindent We also consider the leave-one-out (LOO) GP predictors trained on the set $\mathrm{DoE}_{n-1} := \mathrm{DoE}_{n}\setminus \{(\bm{x}^{(i)},g(\bm{x}^{(i)}))\}$, which can be computed in closed form without retraining the model \citep{Rasmussen2006}. Let $\bm{K}_n^{-1}$ denote the inverse of the covariance matrix and define $\bm{\alpha} = \bm{K}_{n}^{-1}(\bm{g}_n - \beta \bm{1}_{n})$. Then, for $i=1,\ldots,n$, the LOO posterior mean and variance at the training point $\bm{x}^{(i)}$ are given by:
\begin{equation}
\widetilde{g}_{-i}(\bm{x}^{(i)}) = g(\bm{x}^{(i)}) - \frac{\alpha_i}{(\bm{K}_{n}^{-1})_{ii}},\quad \widetilde{\sigma}^2_{-i}(\bm{x}^{(i)}) = \frac{1}{(\bm{K}_{n}^{-1})_{ii}},
\end{equation}
where $\alpha_i$ denotes the $i$-th component of $\bm{\alpha}$ and $(\bm{K}_n^{-1})_{ii}$ the $i$-th diagonal entry of $\bm{K}_n^{-1}$. More generally, for a new input $\bm{x} \in \mathcal{X}$, let $c_i(\bm{x}) := \left[\bm{K}_n^{-1}\bm{k}_n(\bm{x})\right]_i$ denote the $i$-th component of $\bm{K}_n^{-1}\bm{k}_n(\bm{x})$. The LOO predictors can then be expressed as:
\begin{equation}
\widetilde{g}_{-i}(\bm{x}) = \widetilde{g}(\bm{x}) - \frac{c_i(\bm{x})}{(\bm{K}_{n}^{-1})_{ii}} \alpha_i,
\end{equation}
\begin{equation}
\widetilde{\sigma}^2_{-i}(\bm{x}) = \widetilde{\sigma}^2(\bm{x}) + \frac{c_i(\bm{x})^2}{(\bm{K}_n^{-1})_{ii}}.
\end{equation}
\noindent These identities allow efficient computation of LOO quantities required for conformal calibration without explicitly retraining $n$ Gaussian process models. These formulas will be useful for the definitions in the following paragraph. 

\subsection{The J+GP cross-conformal estimator}
\label{sec32}
\noindent The Jackknife+ procedure adapted to GP metamodels \citep{Jaber2025} for constructing adaptive conformal prediction intervals is described in the following. Consider a GP surrogate trained on a dataset $\mathrm{DoE}_{n}$, with hyperparameters $(\beta_{\mathrm{MLE}}, \theta_{\mathrm{MLE}})$ of a certain prior kernel $k_{\theta}$ estimated by maximum likelihood. This model provides the posterior mean $\widetilde{g}$ and standard deviation $\widetilde{\sigma}$ as described in the previous section. We also consider the leave-one-out (LOO) GP models obtained by removing each observation in turn, and denote by $\widetilde{g}_{-i}$ and $\widetilde{\sigma}_{-i}$ the corresponding posterior mean and standard deviation for $i=1,\ldots,n$. We define a normalized LOO non-conformity score that accounts for local predictive uncertainty. Introducing a small constant $\varepsilon > 0$ to avoid degeneracy, we set:
\begin{equation}
    R_{i}^{\,\textnormal{LOO}\sigma}:= \frac{\lvert g(\bm{X}^{(i)}) - \widetilde{g}_{-i}(\bm{X}^{(i)}) \rvert}{\max\left(\varepsilon, \widetilde{\sigma}_{-i}(\bm{X}^{(i)})\right)}, \quad i=1,\ldots,n.
\end{equation}
For a new input $\bm{X}^{(n+1)} \in \mathcal{X}$ and coverage level $1-\alpha \in (0,1)$, we define the J+GP conformal prediction interval as:
\begin{equation}
    \widehat{C}^{\,\textnormal{J+GP}}_{n,\alpha}(\bm{X}^{(n+1)}) =
    \left[ \widehat{q}^{\;\pm}_{n,\alpha}\left\{ \widetilde{g}_{-i}(\bm{X}^{(n+1)}) \pm R_{i}^{\,\textnormal{LOO}\sigma} \times \max\left(\varepsilon, \widetilde{\sigma}_{-i}(\bm{X}^{(n+1)})\right)\right\}\right].
\end{equation}
This construction yields input-dependent intervals whose width adapts to the GP uncertainty. We also introduce a minmax variant, denoted J-minmax-GP, defined by:
\begin{equation}
\begin{aligned}
    \widehat{C}^{\,\textnormal{J-mm-GP}}_{n,\alpha}(\bm{X}^{(n+1)}) =
    \Big[ \min_{i} \widetilde{g}_{-i}(\bm{X}^{(n+1)})
    - \widehat{q}^{\;-}_{n,\alpha}\big\{R_{i}^{\,\textnormal{LOO}\sigma} \times \max(\varepsilon,\widetilde{\sigma}_{-i}(\bm{X}^{(n+1)}))\big\}, \\
    \max_{i} \widetilde{g}_{-i}(\bm{X}^{(n+1)})
    + \widehat{q}^{\;+}_{n,\alpha}\big\{R_{i}^{\,\textnormal{LOO}\sigma} \times \max(\varepsilon, \widetilde{\sigma}_{-i}(\bm{X}^{(n+1)}))\big\}
    \Big].
\end{aligned}
\end{equation}
Both constructions retain the marginal coverage guarantees of their Jackknife+ counterparts \citep{Barber2021} while incorporating GP-based adaptivity \citep{Jaber2025}.
Assume $\mathrm{DoE}_{n}$ is exchangeable. For a new point $\bm{X}^{(n+1)}\in\mathcal{X}$ and a coverage level $1 - \alpha\in(0,1)$, one has:
\begin{equation}
\mathbb{P}_{\mathrm{DoE}_{n+1}}\left(g(\bm{X}^{(n+1)})\in\widehat{C}^{\,\textnormal{J+GP}}_{n,\alpha}(\bm{X}^{(n+1)})\right) \geq1 - 2\alpha.
\end{equation}
and for the minmax variant we get:
\begin{equation}
\mathbb{P}_{\mathrm{DoE}_{n+1}}\left(g(\bm{X}^{(n+1)})\in\widehat{C}^{\,\textnormal{J-mm-GP}}_{n,\alpha}(\bm{X}^{(n+1)})\right) \geq1 - \alpha,
\end{equation}
but in practice these intervals are more conservative and thus could take more time to converge. However, these coverage rates are merely \emph{marginal} meaning on the full set $\mathrm{DoE}_{n+1}$ meaning that it works on average over all the permutations of training and testing datasets. However, in practice we train the model with only a specific design of experiments and it would be more interesting to obtain the so-called \emph{training-conditional} coverage \citep{Angelopoulos2021}. This amounts roughly to replacing $\mathbb{P}_{\mathrm{DoE}_{n+1}}$ by the conditioned probability $\mathbb{P}_{
\bm{X}^{(n+1)},g(\bm{X}^{(n+1)})} (.|\underline{\bm{Z}}_{n} = \mathrm{DoE}_{n} )$ on a \emph{specific} design of experiments. Theoretically this is possible in the cross-conformal cases if the algorithmic model admits certain stability properties \citep{Liang2025}. Such stability for GPs can be obtained under a moderate regularization using a nugget factor (see sec. 4.4 in \citep{Jaber2026-thesis}), thus perturbing slightly the output, this guarantees a stronger training-conditional guarantee:
\begin{equation}
\mathbb{P}_{(\bm{X}^{(n+1)},g(\bm{X}^{(n+1)}))}\left(g(\bm{X}^{(n+1)})\in\widehat{C}^{\,\textnormal{J+GP}}_{n,\alpha}(\bm{X}^{(n+1)} ) \; | \; \mathrm{DoE}_{n} \right) \gtrsim  1 - \alpha.
\end{equation}
However, this probability can be difficult to check and is left for future theoretical work. For now we assume that the obtained marginal coverage property is legitimate. 

\subsection{Active Kriging Monte Carlo Simulation Algorithm}
\label{sec33}
\noindent Let $\mathcal{D} = \{\bm{x}^{(i)}\}_{i=1}^N$ be a Monte Carlo sample drawn from the input distribution $\mathbb{P}_{\bm{X}}$. The goal is to estimate the probability of failure defined in Eq.~\eqref{eq:p_f}
using a GP surrogate and a subset $\mathcal{D}_{*}\subsetneq \mathcal{D}$ such that $|\mathcal{D}_{*}|<<N$. At iteration $t$ of the algorithm, a GP surrogate is trained on a design of experiments $\mathcal{D}_t = \{\bm{x}^{(i)}\}_{i=1}^{n_{t}}$, yielding a posterior mean $\widetilde{g}_t$ and a posterior standard deviation $\widetilde{\sigma}_t$. The Monte Carlo estimate of the failure probability is computed using the surrogate predictor:
\begin{equation}
\widehat{P}_f^{(t)} = \frac{1}{N} \sum_{i=1}^N \mathbf{1}\{\widetilde{g}_{t}(\bm{x}^{(i)}) \leq 0\}.
\end{equation}
The AK-MCS procedure iteratively enriches the design $\mathcal{D}_t$ by selecting a new point $\bm{x}_{t+1}$ from the Monte Carlo sample according to a learning function $\mathcal{L}_t(\bm{x})$, designed to identify the most informative point for improving the estimation of the limit-state surface. The selected point is evaluated using the high-fidelity model $g$, and the surrogate is updated. The algorithm proceeds until a prescribed stopping criterion is satisfied, typically based on the stabilization of the failure probability estimate or on a measure of classification uncertainty.

\subsubsection{U-function}
\label{sec331}
\noindent The classical AK-MCS strategy is based on the so-called U-function \citep{Echard2011}, defined as:
\begin{equation}
U_{t}(\bm{x}) = \frac{|\widetilde{g}_{t}(\bm{x})|}{\widetilde{\sigma}_{t}(\bm{x})}.
\end{equation}
This quantity measures the distance to the limit-state surface in units of posterior uncertainty. The next evaluation point is chosen as:
\begin{equation}
\bm{x}_{t+1} = \argmin_{\bm{x} \in \mathcal{D}} U_{t}(\bm{x}),
\end{equation}
i.e., the point closest to the decision boundary in a standardized Gaussian sense. This strategy aims at refining the surrogate near the estimated limit-state surface and it stops whenever $\min_{\bm{x}\in\mathcal{D}} U_{t}(\bm{x}) \geq 2$ corresponding to roughly $95\%$ of well classified points. However, it is very dependent on the right Gaussian classification of the points near the failure region. 

\subsubsection{Expected Feasibility Function (EFF)}
\label{sec332}

\noindent The Expected Feasibility Function (EFF), originally introduced in the EGRA framework \citep{Bichon2008}, provides a learning criterion that quantifies how well the performance function is expected to satisfy the limit-state condition in a neighborhood of the failure threshold. It favors points that are likely to lie close to the limit-state surface while accounting for the predictive uncertainty of the Gaussian process surrogate. For a given tolerance parameter $\delta > 0$, the EFF measures the expected proximity of $g(\bm{x})$ to the interval $[-\delta, \delta]$ under the Gaussian predictive distribution of the surrogate model. Using the GP predictor $(\widetilde{g}_t(\bm{x}), \widetilde{\sigma}_t(\bm{x}))$, the EFF at iteration $t$ is defined as:
\begin{align}
\mathrm{EFF}_t(\bm{x}) =\; & \widetilde{g}_t(\bm{x}) 
\left[ 2\Phi\!\left( \frac{- \widetilde{g}_t(\bm{x})}{\widetilde{\sigma}_t(\bm{x})} \right)
- \Phi\!\left( \frac{-\delta - \widetilde{g}_t(\bm{x})}{\widetilde{\sigma}_t(\bm{x})} \right)
- \Phi\!\left( \frac{\delta - \widetilde{g}_t(\bm{x})}{\widetilde{\sigma}_t(\bm{x})} \right)
\right] \nonumber \\
& - \widetilde{\sigma}_t(\bm{x})
\left[ 2\phi\!\left( \frac{- \widetilde{g}_t(\bm{x})}{\widetilde{\sigma}_t(\bm{x})} \right)
- \phi\!\left( \frac{-\delta - \widetilde{g}_t(\bm{x})}{\widetilde{\sigma}_t(\bm{x})} \right)
- \phi\!\left( \frac{\delta - \widetilde{g}_t(\bm{x})}{\widetilde{\sigma}_t(\bm{x})} \right)
\right] \nonumber \\
& + \delta \left[
\Phi\!\left( \frac{\delta - \widetilde{g}_t(\bm{x})}{\widetilde{\sigma}_t(\bm{x})} \right)
- \Phi\!\left( \frac{-\delta - \widetilde{g}_t(\bm{x})}{\widetilde{\sigma}_t(\bm{x})} \right)
\right],
\end{align}
where $\Phi$ and $\phi$ denote respectively the cumulative distribution function and the probability density function of the standard normal distribution. Following \citep{Bichon2008}, the tolerance parameter is chosen as $\delta = 2\,\widetilde{\sigma}_t^2(\bm{x})$, ensuring that the exploration region adapts to the local predictive uncertainty. The next enrichment point is selected as:
\begin{equation}
\bm{x}_{t+1} = \argmax_{\bm{x} \in \mathcal{D}} \mathrm{EFF}_t(\bm{x}).
\end{equation}
The usual stopping criterion used is whenever $\max_{\bm{x}\in\mathcal{D}}\mathrm{EFF}_{t}\leq 10^{-3}$ then the algorithm stops \citep{Echard2011}. As we can see, this method too relies heavily on the Gaussian hypothesis of the output. 

\subsubsection{Conformal C$^2$ learning function}
\label{sec333}
\noindent In the proposed AK-MCS-C$^2$ approach, the enrichment is driven by the cross-conformal prediction sets introduced in section \ref{sec32} and does not rely on any model-hypothesis for interpretation. For a given confidence level $1-\alpha$, we consider prediction intervals constructed at iteration $t$ from the design $\mathcal{D}_{t} = \{(\bm{x}^{(i)},g(\bm{x}^{(i)})\}_{i=1}^{n_{t}}$:
\begin{equation}
\widehat{C}^{*}_{n_t,\alpha}(\bm{x}) = [m_{t}(\bm{x}), M_{t}(\bm{x})].
\end{equation}
with $* = \{\mathrm{J+GP},\;\mathrm{J-mm-GP}\}$. Define the set of uncertain points:
\begin{equation}
\mathcal{U}_t = \left\{ \bm{x}\in\mathcal{D} : m_{t}(\bm{x}) \leq 0 \leq M_{t}(\bm{x}) \right\}.
\end{equation}
For $\bm{x} \in \mathcal{U}_t$, we use the diameter function:
\begin{equation}
r_t(\bm{x}) = \frac{1}{2}\mathrm{diam}(\widehat{C}^{*}_{n_t,\alpha}(\bm{x})) = \frac{1}{2}\lvert M_{t}(\bm{x}) - m_{t}(\bm{x})\rvert.
\end{equation}
And the next point is selected as:
\begin{equation}
\bm{x}_{t+1} = \argmax_{\bm{x} \in \mathcal{U}_t} r_t(\bm{x}),
\end{equation}
i.e., the most uncertain point among those whose classification is \emph{ambiguous} with respect to the failure boundary. In addition, the conformal intervals provide bounds on the failure probability:
\begin{equation}
\widehat{P}_f^{+} = \frac{1}{N} \sum_{i=1}^{N} \mathbf{1}\{M_{t}(\bm{x}^{(i)}) \leq 0\},
\qquad
\widehat{P}_f^{-} = \frac{1}{N} \sum_{i=1}^{N} \mathbf{1}\{m_{t}(\bm{x}^{(i)}) \leq 0\},
\label{eq:pf_bounds}
\end{equation}
and we define a stopping criterion similar to \citep{Schobi2017} as:
\begin{equation}
\widehat{P}_f^{+} - \widehat{P}_f^{-} \leq \varepsilon_{\mathrm{stop}},
\label{eq:stopping}
\end{equation}
for a prescribed tolerance $\varepsilon_{\mathrm{stop}} > 0$. \\

\noindent At each iteration $t$ we report the empirical coverage of the conformal intervals, defined as the fraction of the i.i.d Monte Carlo points in $\mathcal{D}$ not yet in the design $\mathcal{D}_{t}$ whose true response falls inside the predicted interval:
\begin{equation}
\widehat{c}_t = \frac{1}{|\mathcal{D}\setminus\mathcal{D}_t|}
\sum_{\bm{x}\in\mathcal{D}\setminus\mathcal{D}_t}
\mathbf{1}\!\left\{ g(\bm{x}) \in \widehat{C}^{*}_{n_t,\alpha}(\bm{x}) \right\}.
\label{eq:emp_cov}
\end{equation}

\section{Numerical results}
\label{sec4}

\subsection{Experimental protocol}
\label{sec40}
\noindent The proposed AK-MCS-C$^2$ method, in both its J+GP and J-minmax-GP variants, is compared against the classical AK-MCS strategies based on the $U$-function (section~\ref{sec331}) and the EFF criterion (section~\ref{sec332}). All learning functions are evaluated under strictly identical conditions: a common Monte Carlo population $\mathcal{D}$ of size $N = 10^{4}$ drawn from a specific $\mathbb{P}_{\bm{X}}$, a common initial design of experiments of size $n_0 = 20$ drawn from $\mathcal{D}$, and the same GP prior (constant trend, anisotropic Mat\'ern-$5/2$ kernel, hyperparameters re-estimated by maximum likelihood at every iteration). The conformal level is fixed to $1 - \alpha = 0.90$ throughout, and the C$^2$ stopping tolerance to $\varepsilon_{\mathrm{stop}} = 10^{-3}$. Each experiment is replicated over $50$ independent random seeds (randomizing both the initial design and the Monte Carlo population), and all figures report the median trajectory together with interquartile shaded bands. Two diagnostics are tracked along the iterations: the failure probability estimate $\widehat{P}_f^{(t)}$ compared to a brute-force reference $P_f^{\mathrm{ref}}$ computed by crude Monte Carlo on the same population, and the empirical coverage of the conformal intervals, i.e.\ the fraction of points of $\mathcal{D}$ (not in the current design) whose true response $g(\bm{x})$ falls inside $\widehat{C}^{*}_{n_t,\alpha}(\bm{x})$. The latter diagnostic is specific to the C$^2$ variants and allows monitoring the validity of the conformal certification during the active-learning process, a quantity which has no analogue for the $U$ and EFF criteria. \\

\noindent It should be emphasized that the empirical coverage of Eq.~\eqref{eq:emp_cov} requires the true responses $g(\bm{x})$ at the unevaluated Monte Carlo points, and is therefore an offline validation tool. It is computable here because the benchmark limit-state functions are analytic. Indeed, it would not be available on a genuinely expensive black-box model except if a held-out evaluation set is available. In a real study, the runtime-computable outputs of the method are the certified bounds $[\widehat{P}_f^{-},\widehat{P}_f^{+}]$ of Eq.~\eqref{eq:pf_bounds} and the stopping rule built on their gap, both of which depend only on the conformal intervals over the Monte Carlo cloud and require no further calls to $g$. The benchmark coverage curves reported below thus serve to establish that this certification remains reliable under adaptive sampling on problems where ground truth is available. All reported terminal values are collected in Table~\ref{tab:summary} below.

\begin{table}[!htb]
\centering
\tiny
\begin{tabular}{llccccc}
\toprule
Benchmark & Method & $\widehat{P}_f$ (median) & Rel.\ err.\ (\%) & CoV (\%) & $n_{\mathrm{calls}}$ & Final coverage \\
\midrule
\multirow{4}{*}{\shortstack[l]{4-branch, $k=6$\\ $P_f^{\mathrm{ref}}=4.46\cdot 10^{-3}$}}
 & U             & $4.45\cdot 10^{-3}$ & $-0.2$  & $18.5$ & $59$ & --- \\
 & EFF           & $4.55\cdot 10^{-3}$ & $+2.0$  & $16.6$ & $51$ & --- \\
 & C$^2$-J+GP    & $4.50\cdot 10^{-3}$ & $+0.9$  & $16.8$ & $58$ & $0.739$ \\
 & C$^2$-J-mm-GP & $4.50\cdot 10^{-3}$ & $+0.9$  & $16.7$ & $66$ & $0.886$ \\
\midrule
\multirow{4}{*}{\shortstack[l]{4-branch, $k=7$\\ $P_f^{\mathrm{ref}}=2.23\cdot 10^{-3}$}}
 & U             & $2.05\cdot 10^{-3}$ & $-8.1$  & $33.2$ & $47$ & --- \\
 & EFF           & $2.15\cdot 10^{-3}$ & $-3.6$  & $23.2$ & $47$ & --- \\
 & C$^2$-J+GP    & $2.15\cdot 10^{-3}$ & $-3.6$  & $22.5$ & $47$ & $0.749$ \\
 & C$^2$-J-mm-GP & $2.10\cdot 10^{-3}$ & $-5.8$  & $23.6$ & $53$ & $0.905$ \\
\midrule
\multirow{4}{*}{\shortstack[l]{Rastrigin\\ $P_f^{\mathrm{ref}}=7.30\cdot 10^{-2}$}}
 & U             & $5.43\cdot 10^{-2}$ & $-25.7$ & $12.8$ & $409$ & --- \\
 & EFF           & $7.35\cdot 10^{-2}$ & $+0.7$  & $3.2$  & $564$ & --- \\
 & C$^2$-J+GP    & $7.35\cdot 10^{-2}$ & $+0.6$  & $12.8$ & $525$ & $0.998$ \\
 & C$^2$-J-mm-GP & $7.28\cdot 10^{-2}$ & $-0.3$  & $10.8$ & $541$ & $0.969$ \\
\midrule
\multirow{4}{*}{\shortstack[l]{Oscillator $6$D\\ $P_f^{\mathrm{ref}}=3.90\cdot 10^{-2}$}}
 & U             & $3.84\cdot 10^{-2}$ & $-1.7$  & $4.4$  & $118$ & --- \\
 & EFF           & $4.12\cdot 10^{-2}$ & $+5.5$  & $9.0$  & $32$  & --- \\
 & C$^2$-J+GP    & $3.84\cdot 10^{-2}$ & $-1.5$  & $4.4$  & $99$  & $0.800$ \\
 & C$^2$-J-mm-GP & $3.84\cdot 10^{-2}$ & $-1.5$  & $4.4$  & $105$ & $0.905$ \\
\bottomrule
\end{tabular}
\caption{Performance summary (median over $50$ replications). Rel.\ err.\ is the median relative error on $\widehat{P}_f$ with respect to $P_f^{\mathrm{ref}}$; CoV is the coefficient of variation of $\widehat{P}_f$ across replications; $n_{\mathrm{calls}}$ is the median number of limit-state evaluations including the $n_0=20$ initial-design points; the final-coverage column reports the median terminal empirical coverage of the conformal intervals (only defined for the C$^2$ variants). The nominal conformal level is $1-\alpha=0.90$.}
\label{tab:summary}
\end{table}

\subsection{$2$D-$4$ branch with $k=6,7$}
\label{sec41}
\noindent The first example consists of a two-dimensional series system with four branches, commonly used as a benchmark problem in structural reliability \citep{Waarts2000,Schueremans2005}. The input random vector $\bm{X} = (X_1, X_2)$ is assumed to follow an independent standard normal distribution $\mathcal{N}(0,1)\otimes\mathcal{N}(0,1)$. The associated limit-state function is defined as:
\begin{equation}
g(\bm{x}) = g(x_1,x_2) = \min \left\{
\begin{aligned}
&3 + 0.1(x_1 - x_2)^2 - (x_1 + x_2)/\sqrt{2}, \\
&3 + 0.1(x_1 - x_2)^2 + (x_1 + x_2)/\sqrt{2}, \\
&(x_1 - x_2) + k/\sqrt{2}, \\
&(x_2 - x_1) + k/\sqrt{2}
\end{aligned}
\right\}.
\end{equation}
The parameter $k$ controls the difficulty of the problem and is set to $k=6$ and $k=7$, following standard configurations in the literature \citep{Echard2011}. The corresponding reference failure probabilities are $P_f^{\mathrm{ref}} = 4.46\times 10^{-3}$ and $P_f^{\mathrm{ref}} = 2.23\times 10^{-3}$, respectively.

\begin{figure}[!htb]
    \centering
    \begin{subfigure}{0.49\linewidth}
        \includegraphics[width=\linewidth]{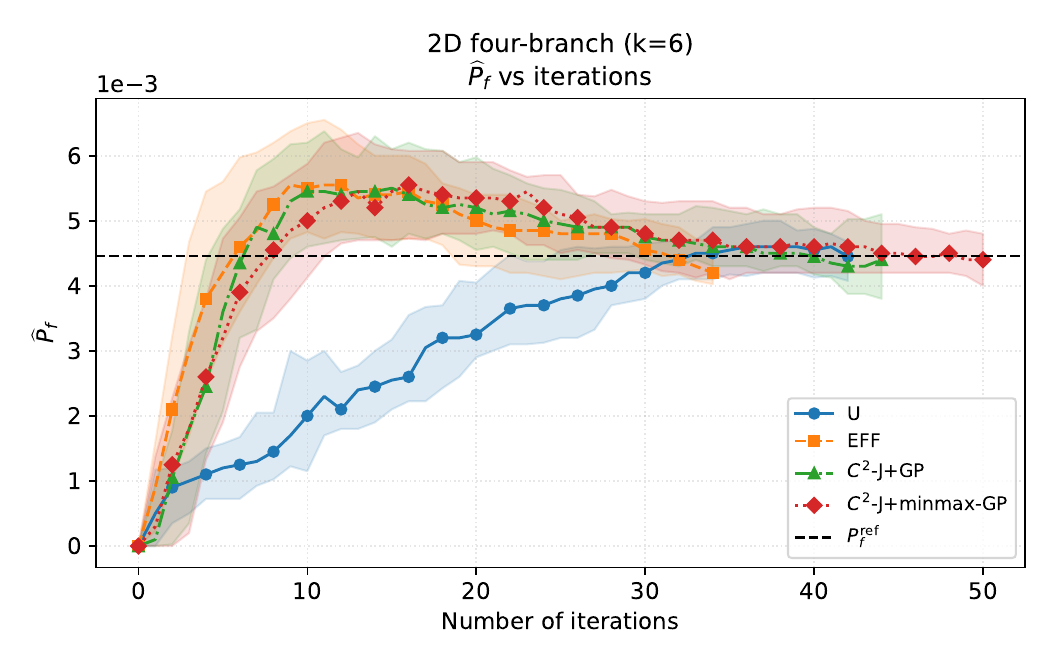}
        \caption{$\widehat{P}_f$ vs.\ iterations, $k=6$.}
    \end{subfigure}
    \begin{subfigure}{0.49\linewidth}
        \includegraphics[width=\linewidth]{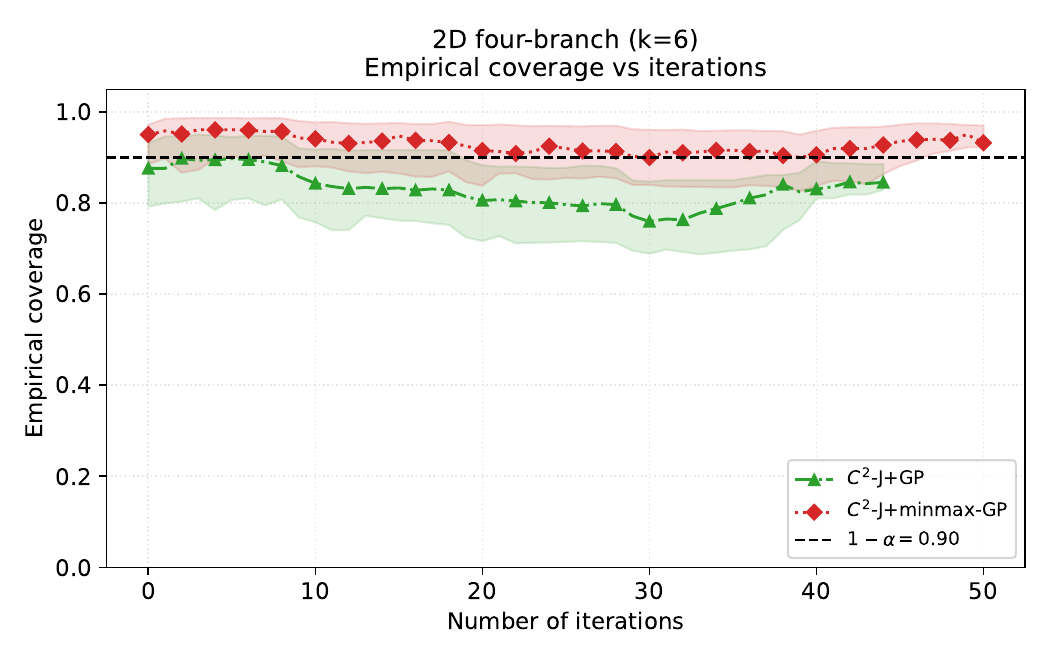}
        \caption{Empirical coverage, $k=6$.}
    \end{subfigure}\\
    \begin{subfigure}{0.49\linewidth}
        \includegraphics[width=\linewidth]{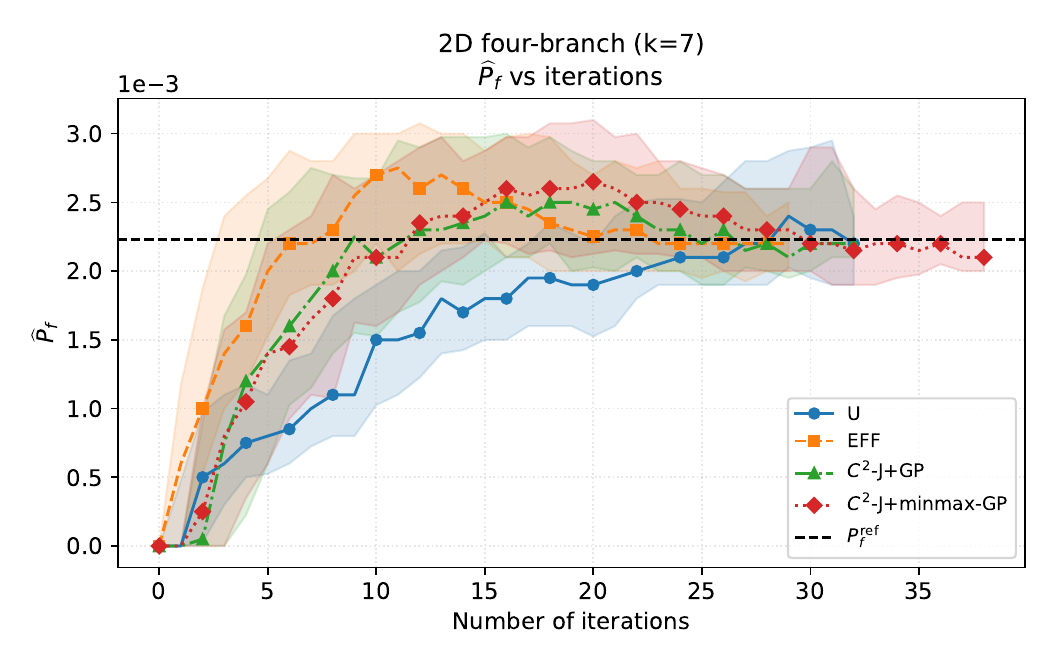}
        \caption{$\widehat{P}_f$ vs.\ iterations, $k=7$.}
    \end{subfigure}
    \begin{subfigure}{0.49\linewidth}
        \includegraphics[width=\linewidth]{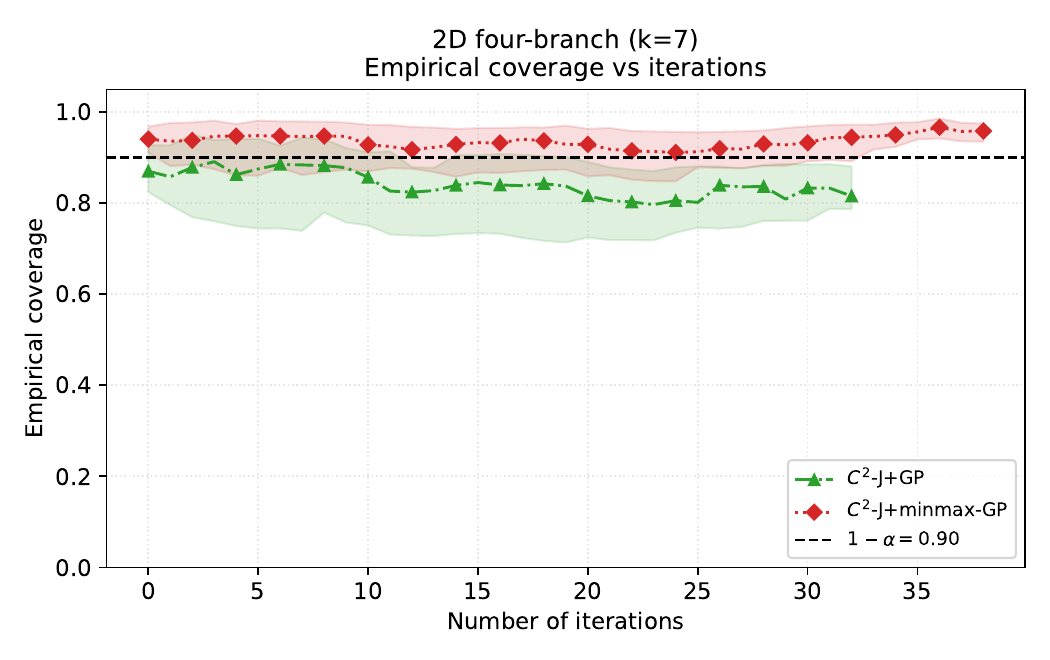}
        \caption{Empirical coverage, $k=7$.}
    \end{subfigure}
    \caption{Two-dimensional four-branch series system. Left: convergence of the failure probability estimate for the four learning functions (median and interquartile bands over $50$ replications). Right: empirical coverage of the conformal intervals on the Monte Carlo population for the two C$^2$ variants, with the nominal level $1-\alpha = 0.90$.}
    \label{fig:branch}
\end{figure}

\noindent Figure~\ref{fig:branch} displays the convergence of $\widehat{P}_f^{(t)}$ and the empirical coverage for both values of $k$. Several observations can be made. First, all strategies reach the reference probability with a similar number of acquisition steps. For $k=6$, EFF and the conformal strategies reach the vicinity of $P_f^{\mathrm{ref}}$ within roughly $8-10$ iterations and stay there, whereas the $U$-criterion approaches the reference from below and reaches it after about $32$ iterations. The same ordering holds for $k=7$. Second, the C$^2$ and EFF trajectories exhibit a transient overshoot of the failure probability (peaking around $5.5\times 10^{-3}$ for $k=6$ near iteration $12$) before relaxing onto the reference. This overshoot is a direct consequence of the exploratory enrichment: wide intervals containing zero in unexplored regions temporarily inflate the estimated failure domain, and the estimate contracts when those regions are resolved. At termination all four methods agree at around $2.0\%$ of $P_f^{\mathrm{ref}}$ for $k=6$ and within $8.1\%$ for $k=7$ (Table~\ref{tab:summary}), at comparable evaluation budgets ($51$-$66$ calls).\\

\noindent The empirical coverage is plotted in the right column. The conservative J-minmax-GP intervals track the nominal $0.90$ level closely throughout the run (median terminal coverage of $0.886$ for $k=6$ and $0.905$ for $k=7$) and their interquartile band remains narrow around the nominal line. The J+GP intervals, which carry the weaker $1-2\alpha$ marginal guarantee, run lower, in the $0.74$-$0.85$ range, as expected. Their coverage dips moderately during the most aggressive enrichment phase (iterations $10$-$30$ for $k=6$) and recovers toward the end as the population near the boundary becomes densely sampled. The systematic ordering J-minmax-GP $\geq$ J+GP, visible at every iteration, is the empirical counterpart of their respective $1-\alpha$ and $1-2\alpha$ guarantees.

\subsection{$2$D modified Rastrigin}
\label{sec42}
\noindent Let $\bm{X} = (X_1,X_2)$ be a random vector with independent standard normal components, i.e.\ $X_i \sim \mathcal{N}(0,1)$. The performance function is defined as:
\begin{equation}
g(\bm{x}) = 10 - \sum_{i=1}^{2} \left( x_i^2 - 5 \cos(2\pi x_i) \right).
\end{equation}
The failure domain is given by:
\begin{equation}
\mathcal{F} = \{ \bm{x} \in \mathbb{R}^2 : g(\bm{x}) \le 0 \}.
\end{equation}
This function exhibits multiple disjoint failure regions and strong nonlinearity, making it a challenging benchmark for reliability analysis \citep{Echard2011}.

\begin{figure}[!htb]
    \centering
    \begin{subfigure}{0.49\linewidth}
        \includegraphics[width=\linewidth]{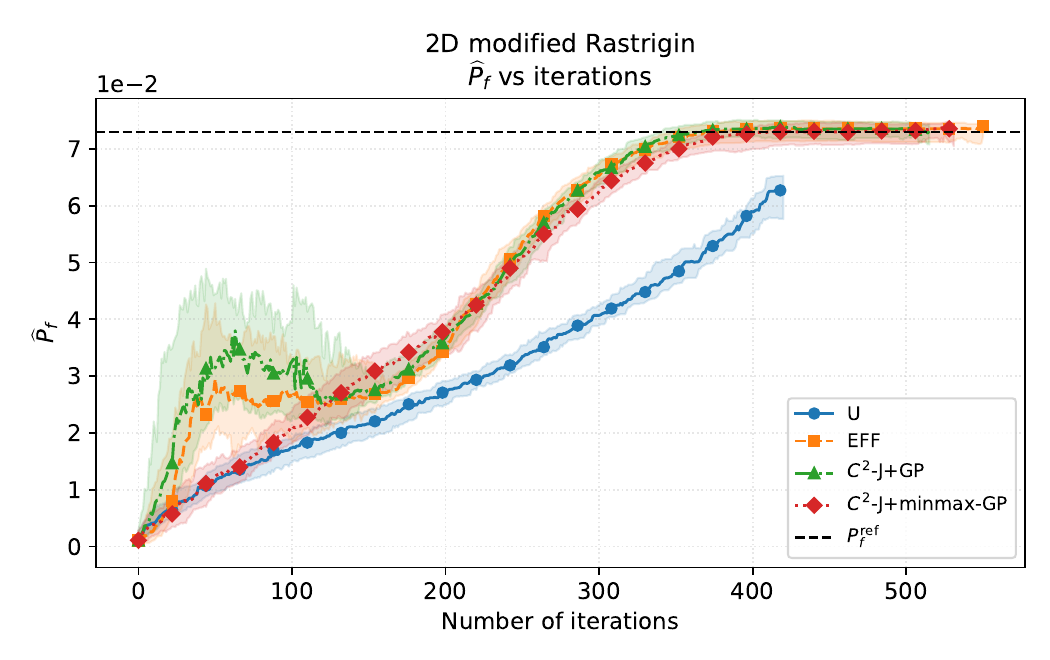}
        \caption{$\widehat{P}_f$ vs.\ iterations.}
    \end{subfigure}
    \begin{subfigure}{0.49\linewidth}
        \includegraphics[width=\linewidth]{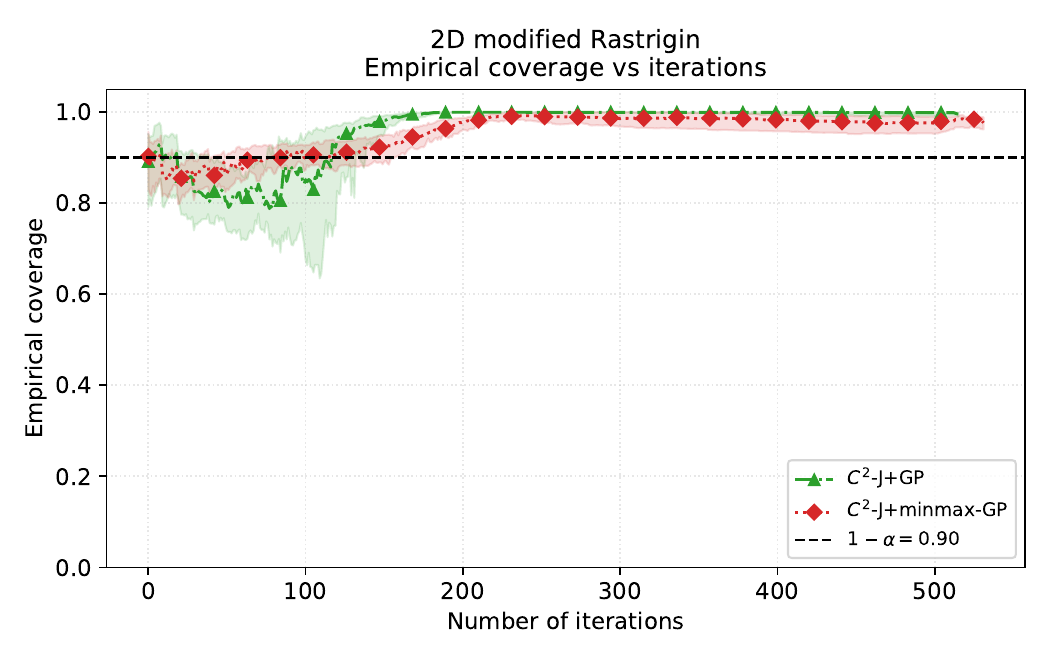}
        \caption{Empirical coverage.}
    \end{subfigure}
    \caption{Two-dimensional modified Rastrigin function. The disconnected, highly multimodal failure domain separates the methods sharply: EFF and both C$^2$ variants recover $P_f^{\mathrm{ref}}$, whereas the $U$-criterion stalls in a severe underestimate.}
    \label{fig:rastrigin}
\end{figure}

\noindent This benchmark discriminates the learning functions more sharply than the preceding four-branch problem. The $U$-criterion performs poorly here: it stalls in a slow, monotone climb and, after $409$ evaluations, terminates at $\widehat{P}_f \approx 5.4\times 10^{-2}$, corresponding to a $-25.7\%$ underestimate of the true reference probability. The mechanism is explained by the multimodality of the domain: the posterior standard deviation collapses inside the explored basins, so the standardized distance $U$ exceeds the stopping threshold while entire failure islands remain undiscovered and confidently misclassified. By contrast, EFF and both C$^2$ variants climb steadily to the reference and stabilize on it by iteration $350$, reaching it within $0.7\%$ (Table~\ref{tab:summary}). It is worth noting that in this test the J+GP variant does not stop prematurely: its non-conformity scores remain large in the under-explored basins, so the uncertain set $\mathcal{U}_t$ keeps adding points there and the certified gap $\widehat{P}_f^{+} - \widehat{P}_f^{-}$ stays above $\varepsilon_{\mathrm{stop}}$ until the failure domain is genuinely resolved. Both C$^2$ variants thus converge with terminal budgets ($525$ and $541$ calls) comparable to EFF ($564$ calls). The coverage panel is also worth pointing out: once the failure islands are progressively discovered (iterations $100$-$200$), the empirical coverage of both variants rises and then stabilizes onto a plateau, with terminal medians of $0.998$ (J+GP) and $0.969$ (J-minmax-GP). We note that such pronounced over-coverage indicates that the intervals are wider than strictly necessary. We conclude that on this disconnected, multimodal limit-state surface the conformal enrichment retains both a correct failure-probability estimate and a valid certification, whereas the $U$-criterion fails to reach the correct result.

\subsection{$6$D nonlinear oscillator}
\label{sec43}
\noindent Consider a single-degree-of-freedom oscillator subjected to a transient excitation \citep{Echard2011,Schobi2017}. The performance function is defined as:
\begin{equation}
g(c_1,c_2,m,r,t_1,F_1)
= 3r - \left|
\frac{2F_1}{m \omega_0^2}
\sin\!\left(\frac{\omega_0^2 t_1}{2}\right)
\right|,
\end{equation}
where:
\begin{equation}
\omega_0 = \sqrt{\frac{c_1 + c_2}{m}}.
\end{equation}
The failure event corresponds to:
\begin{equation}
\mathcal{F} = \{ (c_1,c_2,m,r,t_1,F_1) \in \mathbb{R}^6 : g \le 0 \}.
\end{equation}
The six input variables are assumed independent and normally distributed with supports provided in Table~\ref{tab:oscillator} below.
\begin{table}[!htb]
\centering
\begin{tabular}{|c|c|}
\hline
\textbf{Input variable} & \textbf{Distribution} \\ \hline
$m$ & $\mathcal{N}(1,\,0.05^2)$ \\
$c_1$ & $\mathcal{N}(1,\,0.1^2)$ \\
$c_2$ & $\mathcal{N}(0.1,\,0.01^2)$ \\
$r$ & $\mathcal{N}(0.5,\,0.05^2)$ \\
$F_1$ & $\mathcal{N}(1,\,0.2^2)$ \\
$t_1$ & $\mathcal{N}(1,\,0.2^2)$\\
\hline 
\end{tabular}
\caption{Probabilistic input model of the $6$D oscillator.}
\label{tab:oscillator}
\end{table}
\noindent This example involves a highly nonlinear and implicit dependence on the input variables through the natural frequency $\omega_0$, leading to a complex failure surface in moderate dimension, with $P_f^{\mathrm{ref}} = 3.9\times 10^{-2}$.
\begin{figure}[!htb]
    \centering
    \begin{subfigure}{0.49\linewidth}
        \includegraphics[width=\linewidth]{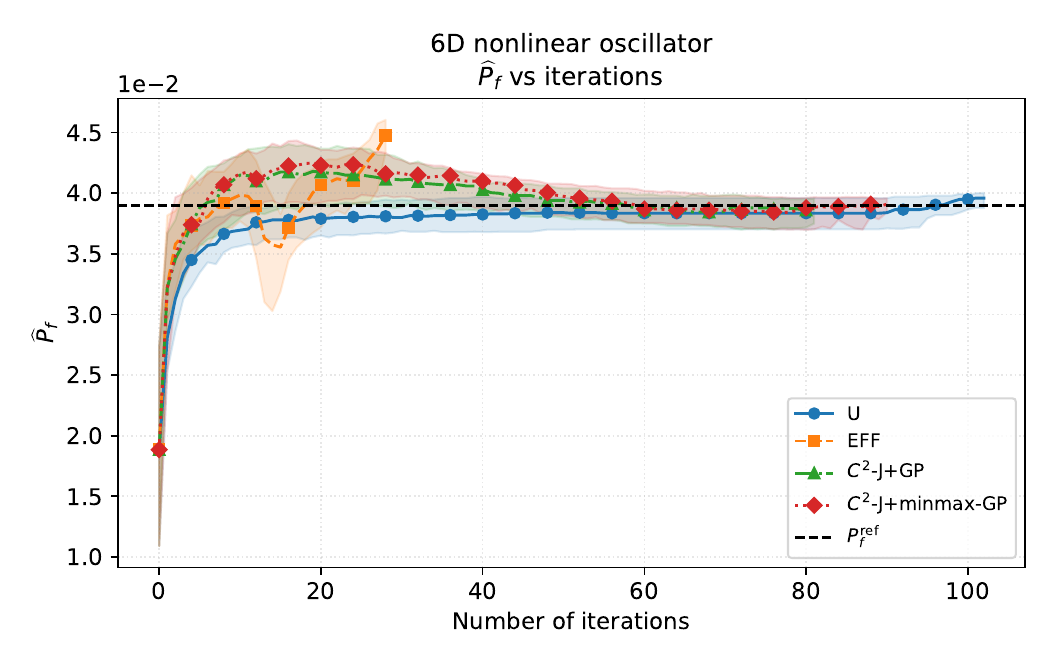}
        \caption{$\widehat{P}_f$ vs.\ iterations.}
    \end{subfigure}
    \begin{subfigure}{0.49\linewidth}
        \includegraphics[width=\linewidth]{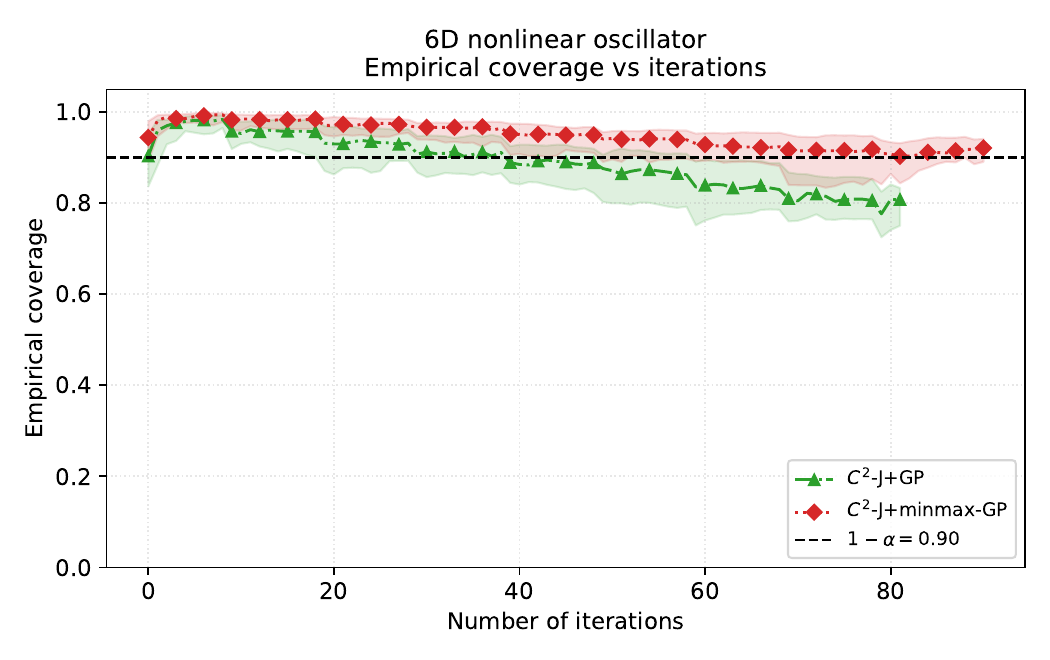}
        \caption{Empirical coverage.}
    \end{subfigure}
    \caption{Six-dimensional nonlinear oscillator. Left: all surrogate-based criteria reach $P_f^{\mathrm{ref}}$, but EFF stops early with a residual bias. Right: the J-minmax-GP coverage stays on the nominal line for the whole run while J+GP erodes mildly as the design concentrates near the limit-state surface.}
    \label{fig:oscillator}
\end{figure}

\noindent In moderate dimension (Figure~\ref{fig:oscillator}), the failure boundary is connected and the $U$-criterion is therefore competitive in accuracy: its Gaussian uncertainty model is approximately correct here, and it converges to within $1.7\%$ of $P_f^{\mathrm{ref}}$, similarly to the two C$^2$ variants ($-1.5\%$ each). The interesting contrast in this example is with the EFF function. The stopping rule is reached early (median $32$ calls) on a transient overshoot, leaving a residual positive bias of $+5.5\%$. The C$^2$ variants instead adjust smoothly onto the reference from above and only stop once the gap closes, at the cost of a larger but more reliable budget ($99$ and $105$ calls). The coverage panel shows the clearest separation between the two conformal constructions. The J-minmax-GP coverage begins slightly above nominal and holds the $0.90$ line for the entire run (terminal median $0.905$), whereas the J+GP coverage starts near nominal and drifts down to approximately $0.80$ as the enrichment concentrates the design closer to the failure boundary. This monotone decrease of the weaker $1-2\alpha$ intervals could point to an empirical signature of exchangeability violation resulting from feedback covariate shift (for more on this see \citep{Fannjiang2022, Stanton2023}). The minmax intervals, being more conservative, absorb that drift and remain valid throughout the learning process. \\

\noindent We stress that the coverage curves of this and the preceding figures are measured against the analytic ground truth and therefore quantify a property that, on a true expensive model, cannot be observed directly. On such a problem one would either set aside a small fraction of the evaluation budget as a held-out calibration check, or transfer the present benchmark evidence as a prior on the reliability of the certification, the certified bounds $[\widehat{P}_f^{-},\widehat{P}_f^{+}]$ themselves remaining computable at every iteration without any additional model evaluation.

\section{Conclusion}
\label{sec5}\noindent This paper introduced AK-MCS-C$^2$, an active-learning reliability method in which the enrichment of the design, the classification of the Monte Carlo sample, the bounding of the failure probability and the stopping decision are all driven by cross-conformal prediction sets rather than by the Gaussian posterior credibility of the Kriging surrogate. It inherits the structure of AK-MCS \citep{Echard2011} while replacing its model-dependent uncertainty measure by the distribution-free, adaptive J+GP and J-minmax-GP intervals of \citep{Jaber2025}, which yield a diameter-based learning function on the ambiguously classified set $\mathcal{U}_t$, certified two-sided bounds $[\widehat{P}_f^{-}, \widehat{P}_f^{+}]$, and a stopping criterion with a clear statistical reading. All results are reproducible with the code at \href{https://github.com/EdgarJaber/AK-MCS-C2}{\texttt{EdgarJaber/AK-MCS-C2}}.\\

\noindent Three conclusions emerge from the experiments. First, the C$^2$ variants are consistently competitive in failure-probability accuracy: they match or beat the $U$-function on every benchmark and avoid the early stopping that biases EFF on the $6$D oscillator. Second, the conservative J-minmax-GP construction is the safest default, never failing to recover $P_f^{\mathrm{ref}}$ and recommended whenever disconnected failure domains cannot be excluded a priori. Third, and unlike any classical AK-MCS criterion, the method yields a certification whose reliability can be assessed: on the benchmarks the J-minmax-GP coverage stays on or above the nominal $0.90$ level throughout, while the J+GP intervals settle on the lower trajectory of their $1-2\alpha$ guarantee, showing that the certification degrades slowly rather than abruptly despite the feedback covariate shift induced by adaptive sampling. This coverage diagnostic requires the true responses and is thus available only on benchmarks. What the practitioner obtains at runtime, at no extra evaluations of $g$, are the certified bounds $[\widehat{P}_f^{-}, \widehat{P}_f^{+}]$ and their stopping rule, whose trustworthiness is precisely what the coverage study establishes. Where the $U$- and EFF-criteria assert reliability only through the Gaussian surrogate hypothesis, AK-MCS-C$^2$ delivers the same estimate together with distribution-free bounds shown to remain empirically valid under the active-learning loop.\\

\noindent Several research directions emerge from this work. A direction of work is related to better understanding training-conditional coverage of cross-conformal estimators under the algorithmic stability using nugget regularization \citep{Jaber2026-thesis}, and the correction of the adaptive-sampling feedback covariate shift via weighted conformal prediction \citep{Tibshirani2019}. On the methodological side, the C$^2$ layer is surrogate-agnostic and could wrap PCE-Kriging for instance \citep{Schobi2017}. Finally, this work paves the way to the development of other conformal-based learning functions for active-learning in structural reliability analysis.

\section*{Acknowledgements}
\noindent This research was conducted as part of a postdoctoral research grant obtained by the first author from the chair of Industrial Data-Analytics \& Machine Learning (IDAML) of Centre Borelli (ENS Paris-Saclay).

\bibliographystyle{elsarticle-num} 
\bibliography{ak-mcs}
\end{document}